\documentclass[format=acmsmall,manuscript,screen,review]{acmart}  %
\usepackage{caption}
\usepackage{hyperref}
\usepackage[utf8]{inputenc}
\usepackage{subcaption}
\usepackage{wrapfig}

\graphicspath{{images/}}
\AtBeginDocument{%
  \providecommand\BibTeX{{%
    \normalfont B\kern-0.5em{\scshape i\kern-0.25em b}\kern-0.8em\TeX}}}

\setcopyright{acmcopyright}
\copyrightyear{2021}
\acmYear{2021}
\acmDOI{10.1145/1122445.1122456}

\begin{document}

\title[UX for COVID-19 News Credibility]{User Experience Design for Automatic Credibility Assessment of News Content About COVID-19}

\author{Konstantin Schulz}
\email{Konstantin.Schulz@dfki.de}
\orcid{0000-0002-3261-9735}
\affiliation{%
  \institution{German Research Center for Artificial Intelligence}
  \streetaddress{Alt-Moabit 91c}
  \city{Berlin}
  \country{Germany}
  \postcode{10559}
}

\author{Jens Rauenbusch}
\author{Jan Fillies}
\email{jfillies@3pc.de}
\author{Lisa Rutenburg}
\author{Dimitrios Karvelas}
\email{dkarvelas@3pc.de}
\affiliation{%
  \institution{3pc GmbH Neue Kommunikation}
  \streetaddress{Prinzessinnenstraße 1}
  \city{Berlin}
  \country{Germany}
  \postcode{10969}
}

\author{Georg Rehm}
\email{georg.rehm@dfki.de}
\affiliation{%
  \institution{German Research Center for Artificial Intelligence}
  \streetaddress{Alt-Moabit 91c}
  \city{Berlin}
  \country{Germany}
  \postcode{10437}
}

%\renewcommand{\shortauthors}{Schulz, et al.}

\begin{abstract}
The increasingly rapid spread of information about COVID-19 on the web calls for automatic measures of quality assurance. In that context, we check the credibility of news content using selected linguistic features. We present two empirical studies to evaluate the usability of graphical interfaces that offer such credibility assessment. In a moderated qualitative interview with six participants, we identify rating scale, sub-criteria and algorithm authorship as important predictors of the usability. A subsequent quantitative online survey with 50 participants reveals a conflict between transparency and conciseness in the interface design, as well as a perceived hierarchy of metadata: the authorship of a news text is more important than the authorship of the credibility algorithm used to assess the content quality. Finally, we make suggestions for future research, such as proactively documenting credibility-related metadata for Natural Language Processing and Language Technology services and establishing an explicit hierarchical taxonomy of usability predictors for automatic credibility assessment.
\end{abstract}




\maketitle

\section{Introduction}
This paper addresses the question of user-centered software design criteria for the automatic assessment of credibility in the COVID-19 domain. We aim in particular to empower citizens in times of a global health crisis, by providing them with relevant, transparent and understandable information about the quality of news content. Formalized quality assurance in information systems has been on a steady rise since at least the year 2000, e.\,g., for news and official speeches \cite{fairbanksCredibilityAssessmentNews2018}. However, it is a challenge for many people, especially younger ones such as students, to distinguish reliably between trustworthy and fake news. They do not reflect critically on the relevant indicators: authorship, primary materials and state of evidence \cite{mcgrewCanStudentsEvaluate2018}. Instead, they are lead astray by mere visual cues of presented content \cite{wobbrockGoldilocksZoneYoung2021}. Their struggle intensifies even more when a distorting political bias in news texts is hidden behind seemingly innocuous category labels in publication organs \cite{kangQuantifyingPerceivedPolitical2020}, or when their own partisanship makes them prejudiced about information quality assessment \cite{saltzMisinformationInterventionsAre2021}. Other groups of people, such as healthcare professionals, are more successful in that respect if topics belong to their domain of expertise, like COVID-19. Even so, they usually do not make use of scientific arguments to explain their reasoning \cite{amitaharonKnowledgeInformationCredibility2021}. As a consequence, our society needs to make high-quality information about critical health issues like COVID-19 more readily available, and help people identify less trustworthy content more easily. Besides, the competent handling of fake news is desirable not only from the perspective of popular education: Expiring domains of fake news websites are often re-registered for criminal purposes, so people returning to their accustomed source of news content are being tricked into various kinds of fraud \cite{chenDiscoveringMeasuringMalicious2021}. The situation gets even more precarious when the general lack of information literacy is combined with a global crisis like COVID-19, leading to an infodemic: rumors and misinformation spread like a disease, making it difficult for people to generate and share reliable knowledge about the existential threat \cite{gallottiAssessingRisksInfodemics2020}. In extreme cases, this exceptionally strong presence of misinformation can be lethal for many people \cite{patwaFightingInfodemicCOVID192021}. There have been proposals for hindering the rapid spread of fake news by applying additional hurdles to the process of information sharing, but they often come at the price of a generally reduced willingness to interact on social platforms \cite{jahanbakhshExploringLightweightInterventions2021}. We therefore have to consider interventions on the receiving end, which usually do not prevent the curation of content. Our focus in this paper is on the end user perspective, which is why we will not address the comparison of different credibility measures or the theoretical distinction of closely related terms like misinformation, disinformation, fake news or trustworthiness. Instead, we seek to develop a design that builds a bridge between well-established research methods and the satisfaction of concrete information needs in the context of COVID-19, which is a known problem in the research community \cite{saltzMisinformationInterventionsAre2021}. In other words, we are more interested in the usability, rather than the functionality, of automatic credibility assessment (ACA) \cite{karrayHumancomputerInteractionOverview2017}.

\section{Related Work}
\subsection{Indicators of Credibility}
Various indicators of credibility have been proposed over the years. Some of them are rather universal, such as wording or preference of content related to single persons, locations or institutions \cite{kakolUnderstandingPredictingWeb2017,fairbanksCredibilityAssessmentNews2018,devernaCoVaxxyCollectionEnglishlanguage2021}. Others are more domain-specific, such as the cite-worthiness and adequate citation of external content \cite{augensteinDeterminingCredibilityScience2021}. Furthermore, the quantifications of each indicator range from simple binary values (`Fake or Real') to advanced multi-dimensional evaluation schemes \cite{suMotivationsMethodsMetrics2020}. This points to a lacking consensus about the definition of some credibility-related concepts, e.\,g., polarity or objectivity \cite{degrandisMultiCriteriaDecisionMaking2019}, leading to inconsistent implementations in practice, sometimes without proper documentation. One important reason for the fuzziness of these conceptualizations is the overlap with other constructs, such as political bias \cite{aksenov2021} or stance, e.\,g., with regard to vaccination against COVID-19 \cite{devernaCoVaxxyCollectionEnglishlanguage2021}. In most cases, this overlap is only partial, which makes many methods effective for assessing one, but not other concepts \cite{fairbanksCredibilityAssessmentNews2018}. Such growing insight into the complexity of credibility assessment has lead to the application of distributed infrastructures \cite{rehm2018c} and entire frameworks like multi-criteria decision making \cite{degrandisMultiCriteriaDecisionMaking2019,pasiDecisionMakingMultiple2020} or ensemble models \cite{dasHeuristicDrivenEnsembleFramework2021} to empower people in their daily struggle against unreliable news content. Furthermore, dedicated institutions like Media Bias Fact Check try to establish themselves as authorities to asses the credibility of online news content \cite{fairbanksCredibilityAssessmentNews2018}.
\label{holisticCredibility}
In many cases, authority is imposed by referring to a multitude of evidence, which can be expressed by a collection of statements or documents supporting a given claim \cite{jiang-etal-2020-hover,thakurBEIRHeterogeneousBenchmark2021}, sometimes condensed into a short explanatory summary \cite{atanasovaGeneratingFactChecking2020a}. Other forms include the aggregation of multiple assessments from independent authorities \cite{rehmInfrastructureEmpoweringInternet2018,zhouReCOVeryMultimodalRepository2020}. This preference of holistic, widespread evidence seems to be in line with common detection patterns for disinformation campaigns (such as Astroturfing), focusing on social coordination instead of individual deviation from a given norm of information handling \cite{kellerPoliticalAstroturfingTwitter2020,vargasDetectionDisinformationCampaign2020}. The specific challenge of integrating multiple aspects and measures of credibility in a single working environment \cite{teyssouInVIDPluginWeb2017} will be discussed further in Section~\ref{uxDesignResearch}.

\subsection{COVID-19 as a Domain for Credibility Assessment}
Depending on the domain, the broad collection of relevant documents has to be accompanied by a certain depth and specialization: If we want to help people recognize fake news about COVID-19, we need to take their information needs into account. What do they want to know? What do they have to know in order to judge a text's credibility reliably? Researchers have made efforts to define relevant topics that contribute to a solid general understanding of COVID-19 \cite{riegerGermanChineseDataset2020}. This domain-specific approach to models of credibility and knowledge suggests itself because recipients of such content are known to be more critical towards user-generated texts if they refer to the medical domain \cite{riehCredibilityAssessmentOnline2014}. Official documents from known authorities such as the government are held in higher esteem. The opposite is true in other domains like tourism: There, people gladly rely on personal experiences and reviews of others that are not tied to specific institutions. Thus, if we want to support individual judgments of content credibility in the COVID-19 domain, we need to provide relevant background information about origin and authorship of a text. As a rule of thumb, we may assume that content distributed through traditional media (newspaper, radio, television) is usually viewed as more credible than texts from the web \cite{riehCredibilityAssessmentOnline2014}. The strong focus on this medical subdomain comes at a price: We lose the ability to create models that are robust and reliable in other domains, which has been noted as a general weakness of research on automated credibility assessment in recent years \cite{suMotivationsMethodsMetrics2020}. On the other hand, we face a more limited object of study, enabling us to build more expressive, more fine-grained representations of knowledge, for instance, in the form of ontologies \cite{duttaCODOOntologyCollection2020,heCIDOCommunitybasedOntology2020} and datasets \cite{sassGermanCoronaConsensus2020,riegerCorona100dGermanlanguageTwitter2021}. Besides, the reduced model complexity allows for easier automation, which is crucial in times of an ever-increasing distribution speed for newly created content \cite{bannonDesignMattersParticipatory2013}: If users can spread their texts in real time, they also need tools for quality assurance in real time, which is hard to achieve without a reasonable amount of computerization \cite{przybylaWhenClassificationAccuracy2021}.

\subsection{User Experience Design}
\label{uxDesignResearch}
When assessing a target group's attitudes towards machine-generated credibility scores, it is advisable to do so in the context of their direct interaction with an associated interface \cite{berndtLearningContextualInquiry2015}. This approach, known as contextual inquiry, offers the benefit of uncovering thoughts that would have gone unnoticed otherwise. Such implicit attitudes are important for judging how well some individuals can satisfy their information needs using that specific interface, which should be seen as the main goal instead of just providing any frontend layer for a given credibility algorithm \cite{gothelfLeanUXDesigning2016}. Ultimately, the outcome of such endeavors should be a service that is provided to the community, not just a software or its source code \cite{juretaComprehensiveQualityModel2009} as required by a number of theoretical user stories \cite{cohnSucceedingAgileSoftware2010,kautzInvestigatingDesignProcess2011,solisStudyCharacteristicsBehaviour2011,wauteletUnifyingExtendingUser2014}. For end users to profit from such a service, it is imperative for them to be involved throughout the design process, thereby becoming software designers themselves \cite{kautzInvestigatingDesignProcess2011}. As a side benefit, the early integration of their feedback and ideas enables the team to react to unforeseen challenges quickly and effectively \cite{leeAgileIntegratedAnalysis2010}. The same goes for software developers: By participating in the design process from the very beginning, they contribute to a cross-functional team that considers problems from multiple perspectives \cite{kuusinenAgileUserExperience2012}, continuing the holistic approach to credibility research outlined in Section~\ref{holisticCredibility}. However, the integration of user experience (UX) design and development work does not necessarily lead to a more streamlined process. Instead, research suggests that designers should usually be `one sprint ahead' of other teams in an agile environment \cite{kuusinenAgileUserExperience2012,raisonKeepingUserCentred2013}, enabling the early anticipation of possible challenges in user interaction. Accordingly, our backend development team used mock objects \cite{samimiDeclarativeMocking2013} to quickly provide Application Programming Interface (API) prototypes while the actual processing logic for ACA was not yet available. When evaluating a given prototype, the tasks and interactions of the chosen users should correspond closely to their everyday behavior, making the study's findings more generalizable \cite{mackenzieHumancomputerInteractionEmpirical2012}.

\subsection{Key Definitions}
\label{sec:keyDefinitions}
For a common understanding and evaluation of the results we are defining the key terms used in our experimental design. This lays the foundation the evaluation of the responses. The definition of transparency has been evolving for many years \cite{michener2013identifying}. In our understanding, we follow Michener \& Bersch \cite{michener2013identifying} who state that transparency is less a theoretical gathering point and rather a descriptive term focusing around state or quality of information. Furthermore, they define that when evaluating transparency, it is sufficient to consider two conditions: visibility and inferability \cite{michener2013identifying}. Understandability can be defined in many ways \cite{houyUnderstandingUnderstandabilityConceptual2012}. We are following the definition of Tu, Tempero \& Thomborson as the degree to which information can be comprehended
with prior knowledge \cite{tu2016experiment}. Relevance has no consensus for a general definition \cite{kagolovskyNewApproachConcept2001}. The underlying problem is that relevance is a temporal and fluid concept that is perceived at a specific moment by a specific user \cite{kagolovskyNewApproachConcept2001}. We are aware of this challenging definition process, but decided to follow Tu, Tempero \& Thomborson and define the term within this study as the degree to which the information obtained by stakeholders answers their questions \cite{tu2016experiment}.

\section{Methodology}
Our algorithm for ACA, called Credibility Score Service, is based on the Credibility Signals\footnote{\url{https://credweb.org/signals-20191126}.} published by the W3C Credible Web Community Group\footnote{\url{https://www.w3.org/community/credibility/}.}. It focuses on the content level and includes linguistic features such as orthography, vocabulary and syntax. Thus, the algorithm does not consider other aspects of credibility such as authorship or distribution platforms, which should be integrated in a separate step as part of future research. We released the source code\footnote{\url{https://github.com/konstantinschulz/alpaca}.} and its documentation\footnote{\url{https://alpaca-credibility.readthedocs.io/en/latest/credibility_signals.html}.} with open licenses to facilitate reuse. Additionally, we offer Docker images\footnote{\url{https://hub.docker.com/r/konstantinschulz/credibility-score-service}.} for enhanced reproducibility and compatibility with various platforms. Finally, we integrated a running instance\footnote{\url{https://live.european-language-grid.eu/catalogue/tool-service/7348}.} of the software into the European Language Grid\footnote{\url{https://www.european-language-grid.eu/}.} \cite{rehmEuropeanLanguageGrid2020,rehm2021a}, a European platform for language technologies. This integration has multiple implications: First of all, it is open and free, which enables us to provide our backend software as a service to not just the design team, but anyone interested in ACA. Thus, the service stays available in a long-term infrastructure, even though the algorithm has originally been developed for the purposes of a time-limited research project, which is a notorious problem in the research community \cite{hettrickResearchSoftwareSustainability2016}. Second, the European Language Grid applies its own metadata management to each of its resources, thereby making them more findable in repositories and, as a consequence, more accessible to a broader public \cite{labropoulouMakingMetadataFit2020a}. Finally, we hypothesize that the involvement of the European Language Grid influences the usability of our credibility service: the location of the servers and, accordingly, the applicability of European laws on data protection (such as 
General Data Protection Regulation, widely known as GDPR) can be important factors for some people when evaluating credibility-related software.

In a human-centered design approach, we developed two graphical user interfaces (GUIs) to visualize results of the Credibility Score Service and receive feedback from potential users. Due to our focus on UX, the limited time frames and budget constraints, we chose to conduct multiple small studies rather than a single larger one. Thus, we were able to evaluate the GUIs in two successive studies and simultaneously make adjustments to the designs. Therefore, the findings from the first survey had an immediate impact on the development of the second GUI. In the following, the design iterations will be referred to as GUI prototypes 1 and 2. 

First, we performed a qualitative evaluation of the GUI prototype 1 by conducting a moderated UX study, which took place remotely with a small number of participants. The aim of this assessment was to gain broad understanding of users' expectations and preferences regarding an online platform aimed at providing information transparency in the context of COVID-19. The evaluation was performed as a formative usability study, i.\,e., it focused on identifying usability problems \cite{sauroQuantifyingUserExperience2016}, as well as an overall assessment of the platform concept. Usability was assessed with regard to interaction design (e.\,g., conforming to GUI conventions), navigation (e.\,g., orientation), visual design (e.\,g., affordance of GUI elements), and wording (e.\,g., user interface copy). The qualitative study consisted of preliminary interviews during which participants are asked about their information-seeking behavior during the pandemic, of a clickable prototype with which the participants perform tasks given by a moderator, and of a short questionnaire they receive after interacting with the prototype. Using a moderated guidance is known to have a positive effect on the overall evaluation of design prototypes \cite{ozencHowSupportDesigners2010}. Throughout the study, participants were encouraged to think aloud. According to Nielsen (1994) \cite{nielsen1994estimating}, a small sample size of four to five participants is sufficient to discover the majority of usability issues in a thinking aloud test. Each of these moderated studies was also simultaneously observed and documented by a second researcher. A description of the prototype can be found in Section~\ref{section:experiments}.

GUI prototype 2 was created after interpreting the findings to address shortcomings of the design, particularly in the realm of information credibility display and interaction. We conducted a remote non-moderated UX survey with a large group of participants to perform quantitative and qualitative evaluation of GUI prototype 2. The aim of the survey was to evaluate the design choices regarding the display of the Credibility Score and the users' interaction with it. The study was evaluated as a summative usability study, i.\,e., it focused on measurements via a survey \cite{sauroQuantifyingUserExperience2016}, and overfulfilled the recommendations (see Budiu \& Moran (2021) \cite{budiu_moran_2021} following Sauro \& Lewis (2016) \cite{sauroQuantifyingUserExperience2016}) of a general sample size of 40 participants for quantitative usability studies. To assess the design choices of the Credibility Score, two key aspects were identified according to the research literature: the origin (e.\,g., intellectual development) and the visual representation of the score (e.\,g., as a scale). The study consists of non-moderated qualitative and non-moderated quantitative parts. Firstly, an understanding of the users' perspective is formed by assessing information-seeking behavior, followed by a short evaluation of the participants' general credibility requirements. After that, an enhanced GUI prototype is presented, along with questions to assess the identified key objectives. All stages use a mix of Likert scales, sliders, open-ended questions and multiple choice questions.

\section{Experiments}
\label{section:experiments}
\subsection{Moderated Remote User Experience Study (GUI Prototype 1)}

\subsubsection{Experiment Setup and Overview of Participants} We performed a moderated remote UX study in July 2021 via video conferencing software and an interactive, web-based prototyping tool. The study was facilitated by a moderator and documented by an observer, who created protocols containing observations and direct quotes from participants. 
Protocols\footnote{\url{https://github.com/konstantinschulz/credible-covid-ux/tree/main/1st-usability-study}.} were anonymized and manually coded using a hybrid approach of inductive and deductive coding \cite{feredayDemonstratingRigorUsing2006}. They will be referred to using a short form (e.\,g., P02, Pos.~61 for participant 2, protocol section 61\footnote{See the file `210702\_Panqura\_Testdesign\_P02.pdf' at \url{https://github.com/konstantinschulz/credible-covid-ux/blob/main/1st-usability-study/210702_Panqura_Testdesign_P02.pdf}.}). The study was conducted in German with six participants living in Germany, from various professional backgrounds (incl. consulting, education, IT, arts, professional services) and across a relatively broad age spectrum from 16 to 69 years (median: 29.5 years). Participants were briefly introduced to the GUI as a new digital platform for information transparency in times of crisis, particularly in the context of the COVID-19 pandemic. It should be noted that in an early stage of the project, the term \textit{trustworthiness} (German: \textit{Vertrauenswürdigkeit}) was used. In prototype 2, it was changed to \textit{credibility} (German: \textit{Glaubwürdigkeit}). However, these terms are closely linked \cite[3]{vivianiCredibilitySocialMedia2017} and we believe that results from the first study can still be used to inform iterations of the GUI prototype. Users were asked to imagine searching for information about the COVID-19 pandemic and vaccines and coming across this information platform which claims to contain up-to-date and transparent information about the subject.

\begin{figure}
    \centering
    \begin{minipage}{.49\textwidth}
      \centering
      \includegraphics[width=\linewidth]{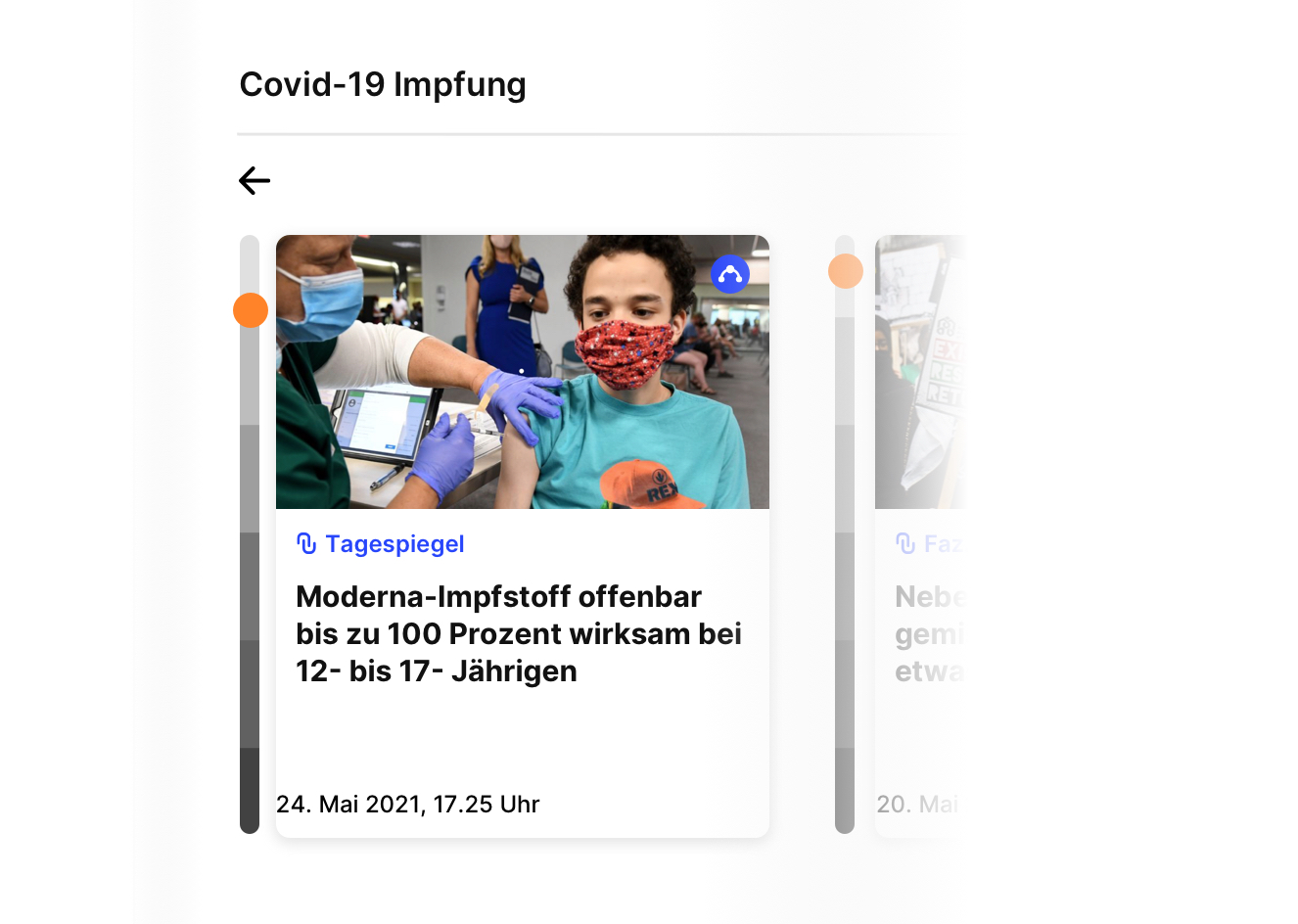}
      \captionof{figure}{Section of the Landing Page in the GUI for Prototype 1.}
      \label{fig:landingPage}
    \end{minipage}
    \hfill
    \begin{minipage}{.49\textwidth}
      \centering
      \includegraphics[width=\linewidth]{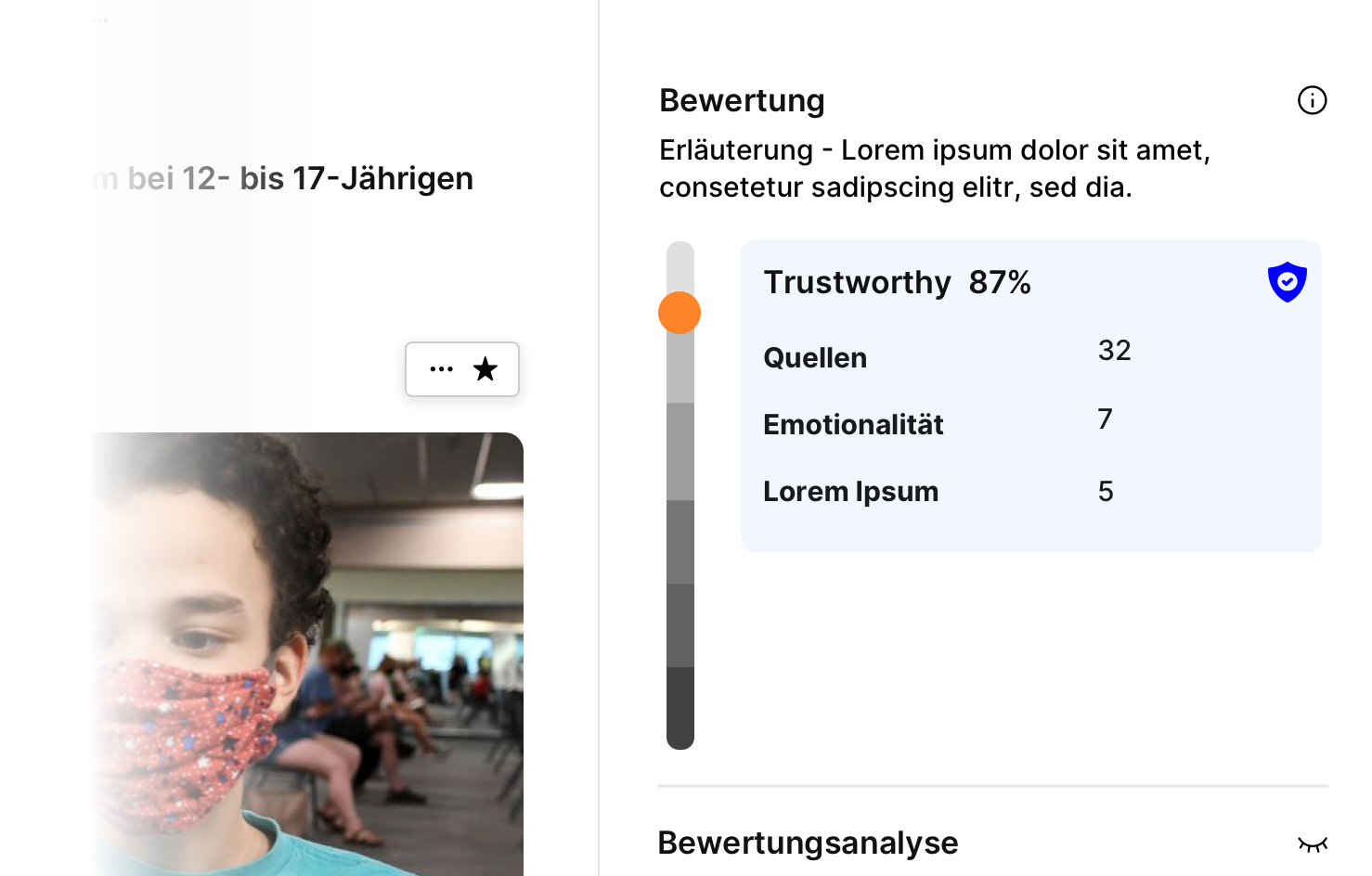}
      \captionof{figure}{Section of the Content View in the GUI for Prototype 1.}
      \label{fig:articlePage}
    \end{minipage}%
\end{figure}

\subsubsection{Description of GUI Prototype 1}
Participants reviewed three screens, i.\,e., sections of the platform. The prototype enabled them to experience a common interaction flow from a landing page\footnote{\url{https://invis.io/WQ11VVW8A79S\#/454763090_Landingpage}.} (dashboard view, see Figure~\ref{fig:landingPage}) to a list of articles on a specific subject (search view) and a detailed display of one specific article (content view, see Figure~\ref{fig:articlePage}).
The dashboard view showed a navigation area (containing the platform's logo and a menu) and three sections on COVID-19-related topics: vaccination, travel regulations, and restrictions in Germany. Each section contained several teasers of news articles which contained article data (image, headline, source, publication date) and additional, analytical information (vertical grey scale visualization indicating an article's trustworthiness, labels representing topics covered by the article). First, users were shown that they could click on the headlines' sections to see more content. Second, users could hover over the grey scale visualization to see a popup with a trustworthiness score, several sub-scores, and a verification icon. By clicking on the headline 
``COVID-19 Vaccination'', participants entered the search view. It contained a prominent headline indicating the topic, a search bar above the headline to filter the results, and a list of search results including an option to change sorting order and a display of the total number of articles. The individual entries, i.\,e., search results, consisted of the same information as the teasers; however, they did not contain images or a trustworthiness score. The GUI also included a sidebar at the right edge of the screen containing various filters. By clicking on the first search result, participants entered the content view, where they could see the news article in full length, including a headline, subhead, an image, and metadata including source, author, and publication date. Additionally, a sidebar at the right edge of the screen displayed the trustworthiness score and sub-scores, an overview table containing metadata on the article, topics, named entities, and placeholders for further analyses. Within the article body itself, several sentences were highlighted in different colors as placeholder visualizations for further analyses.

\subsubsection{Information-seeking Behavior During COVID-19} 
\begin{figure}
    \centering
    \begin{minipage}{.49\textwidth}
      \centering
      \includegraphics[width=\linewidth]{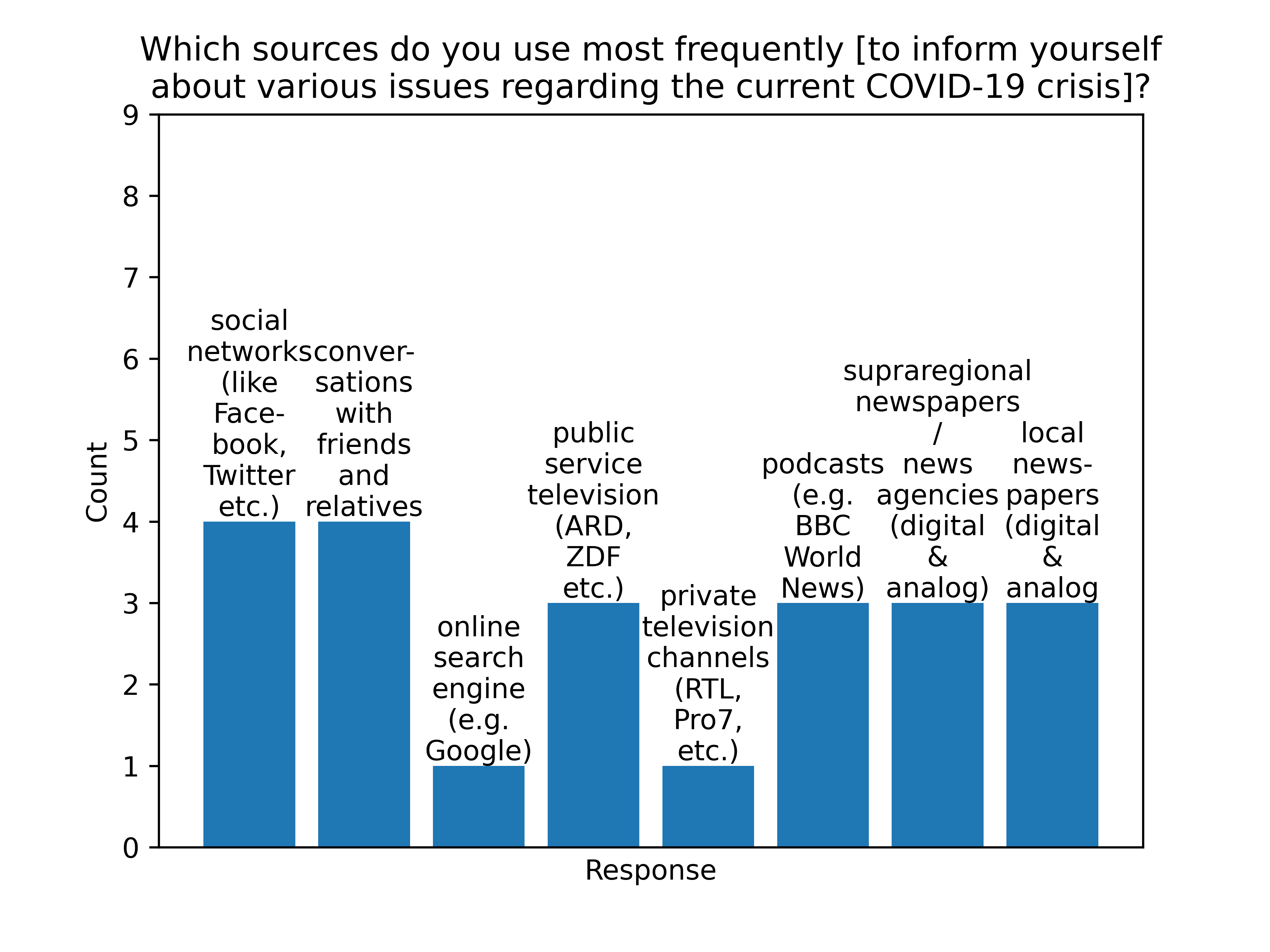}
      \captionof{figure}{Study 1: Response options and counts for the sources used to inform oneself about COVID-19.}
      \label{fig:sourceUsage}
    \end{minipage}
    \hfill
    \begin{minipage}{.49\textwidth}
      \centering
      \includegraphics[width=\linewidth]{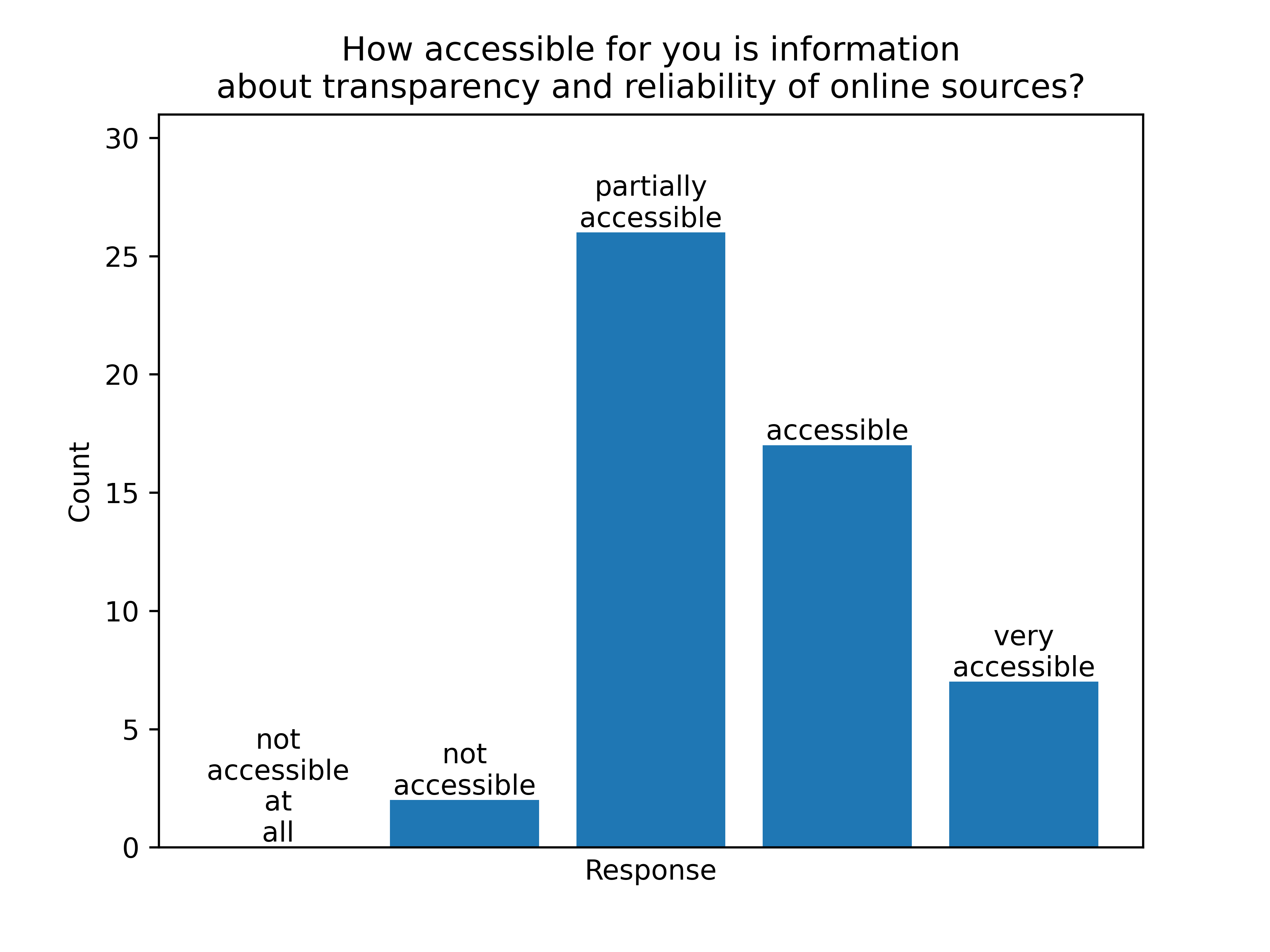}
      \captionof{figure}{Study 2: Response options and counts for the accessibility of information about online sources.}
      \label{fig:Accessibility}
    \end{minipage}%
\end{figure}

During preliminary interviews\footnote{\url{https://github.com/konstantinschulz/credible-covid-ux/blob/main/1st-usability-study/interview_data.csv}.}, most participants stated that they often seek information about the COVID-19 pandemic, i.\,e., daily or several times a day, and mostly do so using social networks, conversations with friends and family, digital news outlets, and public broadcasts (Figure~\ref{fig:sourceUsage}). Participants described these sources as informative, factual, critical, and transparent. They stated that relevant, factual information from transparent and reliable sources were either accessible or very accessible. Participants were mostly concerned with finding insights about current research on COVID-19, current general regulations, political issues, and travel information. Most participants stated that currentness and regionality of information are relevant or very relevant.

\subsubsection{Automatic Credibility Assessment} With regards to ACA, participants were mostly concerned about the source and the scale or interpretation of the credibility score (labeled as \textit{trustworthiness}) while completing tasks regarding GUI prototype 1. Participants repeatedly questioned the source of the automatic assessment. For example, they asked: ``Where does the trustworthiness score come from?''
%\href{https://github.com/konstantinschulz/credible-covid-ux/blob/main/1st-usability-study/210709_Panqura_Testdesign_P05.pdf}{P05, Pos. 61}
(P05, Pos.~61), ``How did [the score] come about?''
%\href{https://github.com/konstantinschulz/credible-covid-ux/blob/main/1st-usability-study/210708_Panqura_Testdesign_P04.pdf}{P04, Pos. 14}
(P04, Pos.~14), and ``Who says that this [news article] is trustworthy?''
%\href{https://github.com/konstantinschulz/credible-covid-ux/blob/main/1st-usability-study/210709_Panqura_Testdesign_P05.pdf}{P05, Pos. 14}
(P05, Pos.~14). Prior to learning more about the automatic assessment, one participant voiced the assumption that the team running the platform evaluates the news articles and performs fact-checking manually
%\href{https://github.com/konstantinschulz/credible-covid-ux/blob/main/1st-usability-study/210702_Panqura_Testdesign_P02.pdf}{P02, Pos. 61}
(P02, Pos.~61). After participants were told that an algorithm automatically assesses trustworthiness, they stated that this information should be made transparent
%\href{https://github.com/konstantinschulz/credible-covid-ux/blob/main/1st-usability-study/210707_Panqura_Testdesign_P03.pdf}{P03, Pos. 101}
(P03, Pos.~101) and that, nevertheless, they still do not understand how decisions are made by the system
%\href{https://github.com/konstantinschulz/credible-covid-ux/blob/main/1st-usability-study/210709_Panqura_Testdesign_P06.pdf}{P06, Pos. 78}
(P06, Pos.~78). Regarding the trustworthiness score itself (including sub-scores such as emotionality), most participants did not understand the scale or how to interpret it. They stated that ``the scale is non-transparent''
%\href{https://github.com/konstantinschulz/credible-covid-ux/blob/main/1st-usability-study/210707_Panqura_Testdesign_P03.pdf}{P03, Pos. 80}
(P03, Pos.~80) and asked ``What is emotionality?''
%\href{https://github.com/konstantinschulz/credible-covid-ux/blob/main/1st-usability-study/210709_Panqura_Testdesign_P06.pdf}{P06, Pos. 15}
(P06, Pos.~15) or ``What does number 7 mean? [Is this the number of] parts which contain emotionality?''
%\href{https://github.com/konstantinschulz/credible-covid-ux/blob/main/1st-usability-study/210702_Panqura_Testdesign_P01.pdf}{P01, Pos. 63}
(P01, Pos.~63). However, some participants stated that they find the scale helpful
%\href{https://github.com/konstantinschulz/credible-covid-ux/blob/main/1st-usability-study/210709_Panqura_Testdesign_P05.pdf}{P05, Pos. 53}
(P05, Pos.~53) or that they would read an article with a low score plainly out of interest and curiosity
%\href{https://github.com/konstantinschulz/credible-covid-ux/blob/main/1st-usability-study/210709_Panqura_Testdesign_P06.pdf}{P06, Pos. 69}
(P06, Pos.~69). When asked whether a trustworthiness score would influence their own assessment of an article, participants' responses were mixed. Some expressed firm belief that such a score would not influence their opinion
%\href{https://github.com/konstantinschulz/credible-covid-ux/blob/main/1st-usability-study/210709_Panqura_Testdesign_P06.pdf}{P06, Pos. 70}
(P06, Pos.~70) and they would only trust it as much as they would understand it
%\href{https://github.com/konstantinschulz/credible-covid-ux/blob/main/1st-usability-study/210709_Panqura_Testdesign_P05.pdf}{P05, Pos. 62}
(P05, Pos.~62). Others expected that a score would indeed influence their opinion and assessment of an article
%\href{https://github.com/konstantinschulz/credible-covid-ux/blob/main/1st-usability-study/210702_Panqura_Testdesign_P01.pdf}{P01, Pos. 68}; \href{https://github.com/konstantinschulz/credible-covid-ux/blob/main/1st-usability-study/210702_Panqura_Testdesign_P02.pdf}{P02, Pos. 67}
(P01, Pos.~68; P02, Pos.~67). Participants frequently stumbled over the visualization of the trustworthiness score, for instance, asking whether they are supposed to click on the mark indicating the score along a grey scale visualization
%\href{https://github.com/konstantinschulz/credible-covid-ux/blob/main/1st-usability-study/210702_Panqura_Testdesign_P02.pdf}{P02, Pos. 15}
(P02, Pos.~15). Furthermore, the verification symbol was associated with verified accounts in popular social networks
%\href{https://github.com/konstantinschulz/credible-covid-ux/blob/main/1st-usability-study/210707_Panqura_Testdesign_P03.pdf}{P03, Pos. 26}
(P03, Pos.~26). The GUI led some participants to believe that the news article had been thoroughly checked for facts
%\href{https://github.com/konstantinschulz/credible-covid-ux/blob/main/1st-usability-study/210708_Panqura_Testdesign_P04.pdf}{P04, Pos. 54}
(P04, Pos.~54), while others stated that they would rather use their own knowledge of online sources to assess their trustworthiness
%\href{https://github.com/konstantinschulz/credible-covid-ux/blob/main/1st-usability-study/210709_Panqura_Testdesign_P06.pdf}{P06, Pos. 39}
(P06, Pos.~39).

\subsubsection{Overall Evaluation of the Prototype} For an overall evaluation of the prototype, participants were asked to rate it in three dimensions using a Likert scale after completing the tasks. Responses were summarized and counted where applicable. The entire platform's \textbf{understandability} was rated as understandable to very understandable and its \textbf{relevance} was rated as relevant to very relevant. The platform's \textbf{transparency} was mostly found to be partly transparent (3x). Other participants rated it as non-transparent (1x), transparent (1x) or very transparent (1x; see the various PDF files).

\subsection{Remote User Experience Survey (GUI Prototype 2)}
\subsubsection{Experiment Setup and Overview of Participants}
We performed a summative usability study in October 2021. In order to generate, publish and find a fitting target group, we used the commercial online platform SurveyMonkey\footnote{\url{https://www.surveymonkey.de}.} and piloted our questionnaire internally.
%a product by Momentive\footnote{\url{https://www.momentive.ai}}
The provided data was anonymously collected. The target audience was compensated and clearly informed about the survey conditions (e.\,g., time frame, success criteria, context). The payment for the participants seemed to be higher than the legal minimum wage, but the exact amount was kept hidden by the survey platform. There was no possibility to interact with the participants during or after the survey. The survey itself was conducted in German. In total, 52 people living in Germany participated in the survey, 50 of which completed the questionnaire. Participants were recruited via the survey platform across all income levels and equally distributed across two genders (female: 25x; male: 25x; the platform did not allow screening for other genders). The rather high age of participants (18-29 years: 12x; 30-44 years: 11x; 45-60 years: 21x; >60 years: 6x) may have induced significant demographic bias \cite{rogersQADatasetExplosion2021} regarding negative attitudes towards artificial intelligence and, thus, ACA \cite{eliasAgeModeratorAttitude2012}. No person below 18 years participated due to legal constraints by the platform. 

\subsubsection{Description of GUI Prototype 2}
\begin{wrapfigure}{r}{0.5\textwidth}
    \begin{center}
      \includegraphics[width=\linewidth]{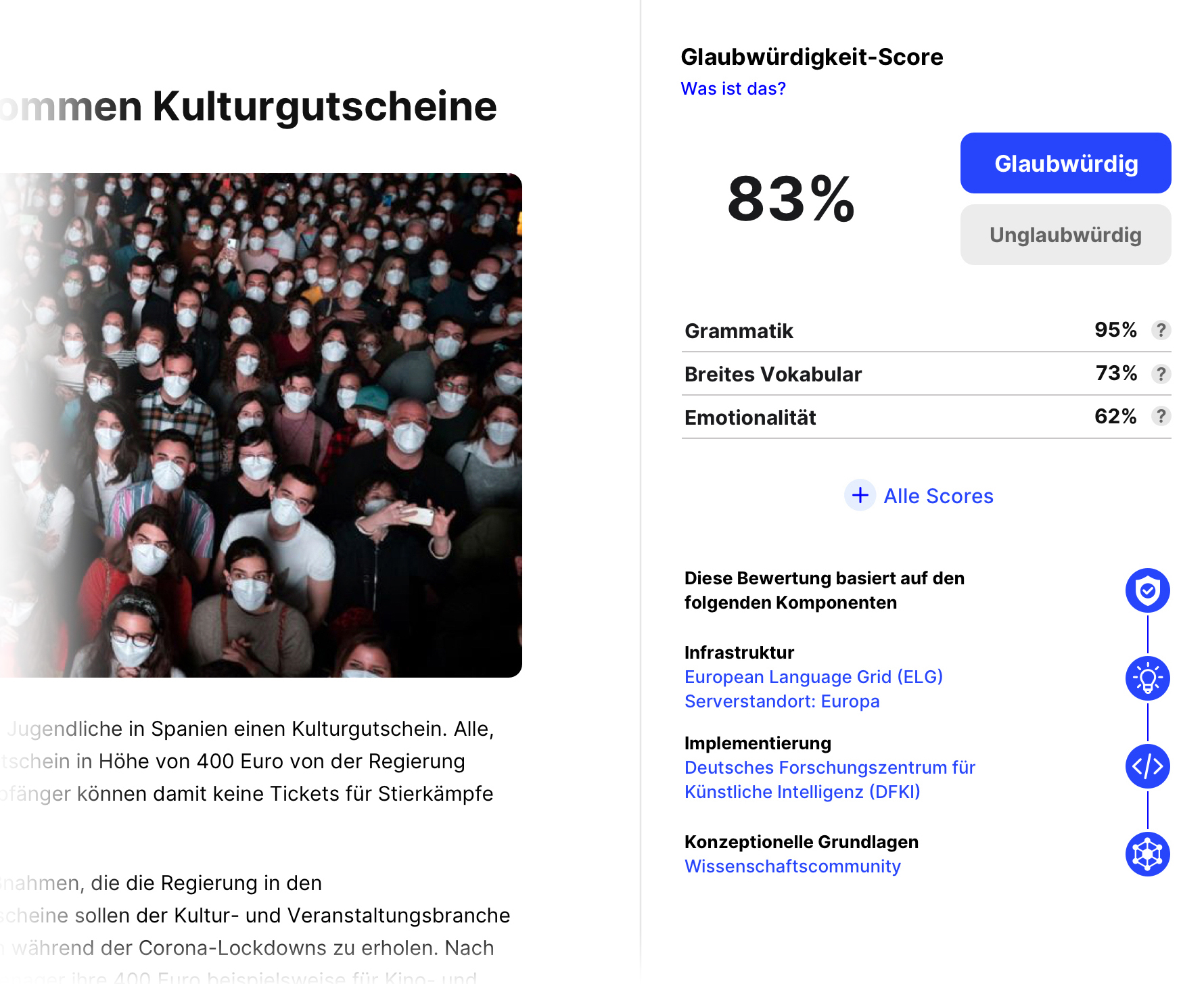}  
    \end{center}
    \caption{Credibility Score focused extract of GUI from Prototype 2.}
    \label{fig:credibilityScore}
\end{wrapfigure}

GUI prototype 2 (see Figure~\ref{fig:credibilityScore}) focused on the credibility score only, thus excluding other sections of the application. It contained an insinuated excerpt of a news article (allowing for larger display of the GUI) and a new section on the right-hand side entitled \textit{Credibility Score}. The new GUI consisted of a fictitious overall score in percent (set to 83\%) and a prominent binary display of credibility classification (credible/incredible; set to credible). Bellow the overall score, three credibility components (grammar, broad vocabulary, emotionality) were displayed with fictitious sub-scores in percentage terms, with the option to view the full list of components with their sub-scores. Furthermore, an overview of authors responsible for creating the score was added, separated into \textit{infrastructure} (``European Language Grid (ELG), Location of Server: Europe''), \textit{implementation} (``German Research Center for Artificial Intelligence (DFKI)''), and \textit{conceptual framework} (``scientific community''). Additionally, the GUI contained elements which suggested that users could receive more information on the overall score and sub-scores. No further metadata about the text (such as author, source, publishing date) was added to the GUI.

\subsubsection{Information-seeking Behavior During COVID-19}
Participants were first asked about their information-seeking behavior during COVID-19. The findings are mostly in line with the qualitative study conducted earlier. Regarding sources of information on the COVID-19 pandemic, more than half (54\%) of the participants stated that they use public television (which may also include consumption of this content via the internet). Slightly less than half use social networks (48\%) and search engines (42\%). Approx.~one third of participants get information through conversations with friends and family, local newspapers, private television (37\% each), radio (33\%), and national newspapers (29\%). 10\% of participants use podcasts and 4\% state that they use none of the sources mentioned. Since multiple answers were possible, the total is greater than 100\%. Similarly to the first study, participants described these sources as informative, factual, critical, and transparent (the less popular choices being: alternative, creative, inviting). One participant added ``convenient''\footnote{\url{https://github.com/konstantinschulz/credible-covid-ux/blob/main/2nd-usability-study/survey_data.csv\#L26}.} as an answer. This illustrates that our curated list of adjectives was successfully geared towards credibility and in line with our research focus. Participants were then asked how accessible they find information about transparency and reliability of online sources. While one half stated it to be only partly accessible, the other half opted for accessible or very accessible, with few exceptions (see Figure~\ref{fig:Accessibility}). It should be noted that while the answers may be subject to central tendency bias, there is still an indication that most people perceived the information as rather accessible.

\subsubsection{Credibility Assessment Criteria}
Participants were introduced to a short news article (see Appendix~\ref{appendix:newspaperArticle}). Then, they were asked to rate the article's credibility and answer related questions. Our goal was to understand their approach to credibility assessment and introduce them to the topic. The article was based on a longer news article by a German news paper\footnote{\url{https://www.spiegel.de/wirtschaft/soziales/post-corona-programm-jugendliche-in-spanien-bekommen-kulturgutscheine-a-d768b360-22dc-473d-afbe-abde5a344482}.}. It consisted of a headline, a subheading, a photo, and three paragraphs of text. No further information (such as author, source or date) was given. Participants rated the article's credibility on a scale from 0\% (not credible at all) to 100\% (completely credible). Ratings varied from 19.5\% (1st quartile) to 66\% (3rd quartile), the median was 49\%. Participants repeatedly mentioned that they mostly consider information about the source, the content itself, and the text's spelling or writing style. Similarly, participants stated that they require more information about the source or author to assess credibility.

\subsubsection{Evaluation of the New Credibility Score GUI}
To assess participants' understanding of the score scale (percentage), they were asked to select an interpretation they believed to be correct. Most participants stated that the credibility score (83\%) should be interpreted as credible or very credible (43\%). Almost equally many participants stated it should be interpreted as partly credible (40\%), which may also be a result of the central tendency bias. Only 17\% stated it should be interpreted as (very) incredible. We have to take into account that question 4\footnote{See \url{https://github.com/konstantinschulz/credible-covid-ux/blob/main/2nd-usability-study/Data_All_211020-broad-audience.pdf}, p. 5.}, in which participants were asked to assess the article's credibility, had already introduced the scale and likely influenced its interpretation by the participants. Furthermore, to evaluate participants' understanding of the score's source, they were asked to state who they believed to be the originator of the credibility score. Some participants stated that they did not know who created the score. However, a pattern emerged in which most participants mentioned the three organizations shown in the lower section of the GUI, particularly the organization mentioned under the heading \textit{implementation}, i.\,e., German Research Center for Artificial Intelligence. The question likely directed participants' attention towards the source, which they might not have noted otherwise. Participants were asked to rate GUI prototype 2 using a Likert scale in the same three dimensions used to assess prototype 1. However, the questions were phrased slightly differently in both studies. Besides, the second study was focused on the credibility score, whereas the first study contained the entire application, including sections which may not be directly related to credibility.

The new prototype was mostly perceived as partly \textbf{understandable}, with a tendency towards understandable (partly understandable: 42\%; (very) understandable: 36\%; not understandable (at all): 22\%).  It was rated as neither extremely \textbf{relevant} nor irrelevant (partly relevant: 46\%; (very) irrelevant: 28\%; (very) relevant: 26\%). Similarly, the new prototype was rated as partly \textbf{transparent} by almost half of all participants (46\%), while the other half was almost equally distributed to (very) non-transparent and (very) transparent, showing a slight tendency towards non-transparent (see questions 7-9). When asked whether the credibility score (83\%) influenced their own assessment of the news article's credibility (median of 49\%), most participants claimed to be uninfluenced. Among the other answers, a slight tendency towards higher credibility (as opposed to lower credibility) was visible (see question 10). Finally, participants were asked to add any further comments or questions. Some participants stated that they still do not know how exactly the score was calculated (``there should be a better explanation for non-professionals how the score is calculated''). One participant stated regarding the sub-scores that ``correct grammar does not reveal anything [about credibility]''. Apart from this comment, sub-scores were not mentioned by the participants.

\section{Results and Discussion}
\subsection{Automatic Credibility Assessment}
Participants' understanding of the credibility score's scale apparently improved through design iterations. In the first study, participants had many questions about how to interpret the scale of the main score, while in the second study, there were few to no questions. Furthermore, participants' answers on how to interpret the scale in the second study showed a tendency towards our intended interpretation, even though there were still some deviations (besides the central tendency bias). Possible explanations include our improved indicators of whether a score should be interpreted as credible or incredible, using a binary display of these categories. Besides, participants in the second study rated the article's credibility before viewing the score, thereby being introduced to the scale. Furthermore, participants of the second study barely mentioned the sub-scores, whereas the first prototype raised many questions in this regard. This may be due to design improvements or changes in the experimental setup, or both. The sub-scores in the second prototype have more specific wording and consistent percentages, which may have caused less irritation among participants. At the same time, these changes did not improve the overall perceived transparency of our ACA. We infer that a simplified textual display of the score's meaning (i.\,e., a relationship between the percentage and terms credible/incredible) helps users comprehend the scale and should also be considered for other services which return a numerical value. Furthermore, manually completing the assessment once before interacting with ACA may help users understand (and critically reflect on) ACA. Future research may also specifically focus on perception and understanding of sub-scores, which play a crucial role in the conceptualization of the main score, and may be conducted via, e.\,g., eye tracking.

\subsection{Algorithm Transparency, Understandability, and Relevance}
In the first study, participants repeatedly asked about the source of the ACA (missing in the GUI). In response, the iterated prototype pointed participants of the second survey explicitly towards an elaborate display of the algorithm's source. Still, many participants stated that they did not know who created the score and the changes did not improve overall transparency, understandability, or relevance. Nevertheless, our qualitative analysis of free text comments reveals that participants were much more upset about the missing article source. This leads to the hypothesis that the source of content is more important than the source of ACA. In other words, there may be a stage-gate process of assessing credibility, where users first require several pieces of basic information (such as source and author of the content) before considering ACA. A significant share of participants, however, directly referred to GUI for the source of the ACA. Among them, some participants stated all three organizations, while others only stated one or two or paraphrased the source. The organization responsible for algorithm implementation was mentioned more often than those responsible for infrastructure or conceptual foundations. Users seem to attribute the highest responsibility to the implementation, perhaps due to association of the term \textit{implementation} with artificial intelligence, or computers in general. Or maybe all three terms (infrastructure, implementation, conceptual foundations) were too vague for users to make informed statements about the source, leading them to pick the entity which is perceived as the most scientific or authoritative. Overall, participants in both studies asked for more contextual information on both content and algorithm. This is a common theme in UX studies: Users request more written explanations in the GUI, for instance via help buttons and popup windows.
%add reference?
However, field studies and eye tracking often reveal that users ignore additional information such as long UX copy. There is a general trade-off in the GUI between informativity or transparency on the one side, and conciseness, clarity, and ease of use on the other. A carefully constructed balance is needed to provide ACA to users effectively. Our findings are helpful, but strongly indicate that our designs have not yet achieved the desired trade-off.

\subsection{Reflections on Methodology}
We noticed a strong central tendency bias in many of the questions, which is a known problem in surveys with scalar response options where a medium value can be chosen \cite{crosettoCentralTendencyBias2020}. This effect intensifies even more for questions with a high cognitive load \cite{allredWorkingMemorySpatial2016}, as was evident in our inquiries about understandability, relevance and transparency. Furthermore, participants' ratings of the GUI slightly worsened in the dimensions of understandability and relevance. While this may be attributed to changes in the design, there are several indicators that these declines are rather based on a change of methodology. First, the remote UX study was performed as a qualitative and moderated assessment in which participants engaged in face-to-face conversations with researchers. This may lead to more positive (or less harsh) responses due to social desirability bias \cite{williams1977experimental}. Second, the subject of evaluation changed throughout the studies. In the first study, participants were asked to rate the overall prototype, including sections which were not related to ACA. In the second study, they were asked to specifically rate the credibility score GUI. As mentioned previously, this research took an iterative, agile approach in which several smaller studies with less preparation time and budget were conducted, as opposed to one larger study. Therefore, the design of both studies likely influenced the results in various ways that need to be considered when interpreting the findings. For example, viewing the rather high score of the ACA in the second study after manually assessing credibility of the article may have led some participants to state that the score increased their perceived credibility of the article. We did not evaluate whether a different order of questions or a lower score in the GUI (i.\,e., lower than 49\%) would have reversed this effect. 

\section{Conclusion}
In the second study, participants seemed ambivalent about the article's credibility, which did not have any contextual information but could only be judged based on the content. Adding the ACA did not really change that. Since the score was generally interpreted according to our intentions, the assumption is that either users heavily rely on contextual information (metadata such as author, source, publishing date) to assess credibility and noted its absence in the GUI; or the score is lacking information which makes it transparent – or both. Regarding the latter, adding information about the creators of the score did not bring the desired effect of adding transparency and users still had questions. As a takeaway, more attention should be paid to providing metadata for the content, as well as general benchmarks for average credibility in the news domain. In some cases, participants admitted to being influenced by the given credibility score. Others denied it, criticizing the underlying criteria and questioning the authority of the people and institutions behind the algorithm. This shows that automatic assessments can provide helpful guidance for end users, but only if they agree with the conceptual basis of the measurements and if they trust the providers of the score. Unfortunately, authorship in terms of software as a service is a fuzzy concept for many people: Some cite the scientific community, others the software developers, still others refer to the server infrastructure while trying to attribute responsibility for a given calculated result. For most people, the source of an algorithm is less important than the authorship of the actual text when it comes to ultimately deciding on its credibility. This puts our efforts to make computational models more explainable into a different perspective: In the future, we should aim to establish a hierarchy of desirable information for assessing content credibility.

Besides, we want to adapt the Credibility Score Service according to our insights from the described studies. In particular, since the transparency and understandability of the score and its components did not reach a sufficient level, the API needs to be modified. Instead of just providing a label and a value for each component, it should also include short descriptions in its metadata (or in responses to invocations), briefly explaining the computation and meaning of each part of the score. Besides, due to the continuing demand for information about origin and authorship, the service should be associated with metadata about its development and infrastructure:
\begin{itemize}
    \item Who invented the concept of ACA?
    \item Who implemented this particular credibility measure?
    \item Who runs this service and where are the servers located? Which data protection laws apply?
\end{itemize}
Finally, we are aiming to expand the basis of the Credibility Score Service by integrating and evaluating information about content authorship, dissemination platforms, coordination patterns and external fact-checking platforms.

\begin{acks}
The research presented in this paper is funded by the German Federal Ministry of Education and Research (BMBF) through the project PANQURA (\url{http://qurator.ai/panqura}; grant no. 03COV03E).

%This work has been developed in the project \href{https://qurator.ai/panqura/}{PanQura}. PanQura (reference number: 03COV03E) is partly funded by the German ministry of education and research (BMBF) within the research programme ICT 2020.

We are grateful to Yuewen Röder (3pc GmbH Neue Kommunikation, Germany) for assisting in the research; to León Viktor Avilés Podgurski for his research on and implementation of credibility signals.
\end{acks}

\bibliographystyle{ACM-Reference-Format}
\bibliography{literature}

\begin{thebibliography}{10}
\providecommand{\url}[1]{\texttt{#1}}
\providecommand{\urlprefix}{URL }
\providecommand{\doi}[1]{https://doi.org/#1}

\bibitem{aksenov2021}
Aksenov, D., Bourgonje, P., Zaczynska, K., Ostendorff, M., Moreno-Schneider,
  J., Rehm, G.: {Fine-grained Classification of Political Bias in German News:
  A Data Set and Initial Experiments}. In: Mostafazadeh~Davani, A., Kiela, D.,
  Lambert, M., Vidgen, B., Prabhakaran, V., Waseem, Z. (eds.) Proceedings of
  the 5th Workshop on Online Abuse and Harms (WOAH~2021). pp. 121--131.
  Association for Computational Linguistics (ACL), Bangkok, Thailand (8 2021),
  co-located with ACL-IJCNLP~2021. 1-6 August 2021

\bibitem{allredWorkingMemorySpatial2016}
Allred, S.R., Crawford, L.E., Duffy, S., Smith, J.: Working memory and spatial
  judgments: Cognitive load increases the central tendency bias. Psychonomic
  Bulletin \& Review  \textbf{23}(6),  1825--1831 (Dec 2016)

\bibitem{amitaharonKnowledgeInformationCredibility2021}
Amit~Aharon, A., Ruban, A., Dubovi, I.: Knowledge and information credibility
  evaluation strategies regarding {{COVID}}-19: A cross-sectional study.
  Nursing Outlook  \textbf{69}(1),  22--31 (2021)

\bibitem{atanasovaGeneratingFactChecking2020a}
Atanasova, P., Simonsen, J.G., Lioma, C., Augenstein, I.: Generating {{Fact
  Checking Explanations}}. In: Proceedings of the 58th {{Annual Meeting}} of
  the {{Association}} for {{Computational Linguistics}}. pp. 7352--7364.
  {Association for Computational Linguistics}, {Online} (Jul 2020).
  \doi{10.18653/v1/2020.acl-main.656}

\bibitem{augensteinDeterminingCredibilityScience2021}
Augenstein, I.: Determining the {{Credibility}} of {{Science Communication}}.
  In: Proceedings of the {{Second Workshop}} on {{Scholarly Document
  Processing}}. pp.~1--6. {Association for Computational Linguistics}, {Online}
  (Jun 2021)

\bibitem{bannonDesignMattersParticipatory2013}
Bannon, L.J., Ehn, P.: Design {{Matters}} in {{Participatory Design}}. In:
  Simonsen, J., Robertson, T. (eds.) Routledge International Handbook of
  Participatory Design, vol.~711, pp. 37--63. {Routledge}, {London \& New York}
  (2013)

\bibitem{berndtLearningContextualInquiry2015}
Berndt, E., Furniss, D., Blandford, A.: Learning {{Contextual Inquiry}} and
  {{Distributed Cognition}}: A case study on technology use in anaesthesia.
  Cognition, Technology \& Work  \textbf{17}(3),  431--449 (Aug 2015)

\bibitem{budiu_moran_2021}
Budiu, R., Moran, K.: How many participants for quantitative usability studies:
  A summary of sample-size recommendations (Jul 2021),
  \url{https://www.nngroup.com/articles/summary-quant-sample-sizes/}

\bibitem{chenDiscoveringMeasuringMalicious2021}
Chen, Z., Freire, J.: Discovering and {{Measuring Malicious URL Redirection
  Campaigns}} from {{Fake News Domains}}. In: 2021 {{IEEE Security}} and
  {{Privacy Workshops}} ({{SPW}}). pp.~1--6. {IEEE}, {San Francisco} (2021)

\bibitem{cohnSucceedingAgileSoftware2010}
Cohn, M.: Succeeding with {{Agile}}: Software {{Development Using Scrum}}.
  {Pearson Education}, {Ann Arbor} (2010)

\bibitem{connordesaiComparingUseOpen2019}
Connor~Desai, S., Reimers, S.: Comparing the use of open and closed questions
  for {{Web-based}} measures of the continued-influence effect. Behavior
  Research Methods  \textbf{51}(3),  1426--1440 (Jun 2019).
  \doi{10.3758/s13428-018-1066-z}

\bibitem{crosettoCentralTendencyBias2020}
Crosetto, P., Filippin, A., Katu{\v s}{\v c}{\'a}k, P., Smith, J.: Central
  tendency bias in belief elicitation. Journal of Economic Psychology
  \textbf{78},  102273 (2020)

\bibitem{dasHeuristicDrivenEnsembleFramework2021}
Das, S.D., Basak, A., Dutta, S.: A {{Heuristic}}-{{Driven Ensemble Framework}}
  for {{COVID}}-19 {{Fake News Detection}}. In: Chakraborty, T., Shu, K.,
  Bernard, H.R., Liu, H., Akhtar, M.S. (eds.) Combating {{Online Hostile
  Posts}} in {{Regional Languages}} during {{Emergency Situation}}. pp.
  164--176. Communications in {{Computer}} and {{Information Science}},
  {Springer International Publishing}, {Cham} (2021).
  \doi{10.1007/978-3-030-73696-5_16}

\bibitem{degrandisMultiCriteriaDecisionMaking2019}
De~Grandis, M., Pasi, G., Viviani, M.: Multi-{{Criteria Decision Making}} and
  {{Supervised Learning}} for {{Fake News Detection}} in {{Microblogging}}. In:
  Workshop on {{Reducing Online Misinformation Exposure}}. pp.~1--8. {ACM},
  {Paris, France} (Jul 2019)

\bibitem{devernaCoVaxxyCollectionEnglishlanguage2021}
DeVerna, M.R., Pierri, F., Truong, B.T., Bollenbacher, J., Axelrod, D., Loynes,
  N., {Torres-Lugo}, C., Yang, K.C., Menczer, F., Bryden, J.: {{CoVaxxy}}: A
  collection of {{English}}-language {{Twitter}} posts about {{COVID}}-19
  vaccines. In: Proceedings of the {{Fifteenth International AAAI Conference}}
  on {{Web}} and {{Social Media}} ({{ICWSM}} 2021). pp. 992--999. {AAAI},
  {Virtual} (2021)

\bibitem{duttaCODOOntologyCollection2020}
Dutta, B., DeBellis, M.: {{CODO}}: An {{Ontology}} for {{Collection}} and
  {{Analysis}} of {{Covid}}-19 {{Data}}. In: Proceedings of the 12th
  {{International Joint Conference}} on {{Knowledge Discovery}}, {{Knowledge
  Engineering}} and {{Knowledge Management}}. pp. 76--85. {SCITEPRESS - Science
  and Technology Publications}, {Budapest, Hungary} (2020).
  \doi{10.5220/0010112500760085}

\bibitem{eliasAgeModeratorAttitude2012}
Elias, S.M., Smith, W.L., Barney, C.E.: Age as a moderator of attitude towards
  technology in the workplace: Work motivation and overall job satisfaction.
  Behaviour \& Information Technology  \textbf{31}(5),  453--467 (2012)

\bibitem{fairbanksCredibilityAssessmentNews2018}
Fairbanks, J., Fitch, N., Knauf, N., Briscoe, E.: Credibility assessment in the
  news: Do we need to read? In: Proc. of the {{MIS2 Workshop}} Held in
  Conjunction with 11th {{Int}}'l {{Conf}}. on {{Web Search}} and {{Data
  Mining}}. pp.~1--8. {ACM}, {Marina Del Rey} (2018)

\bibitem{feredayDemonstratingRigorUsing2006}
Fereday, J., {Muir-Cochrane}, E.: Demonstrating rigor using thematic analysis:
  A hybrid approach of inductive and deductive coding and theme development.
  International journal of qualitative methods  \textbf{5}(1),  80--92 (2006)

\bibitem{gallottiAssessingRisksInfodemics2020}
Gallotti, R., Valle, F., Castaldo, N., Sacco, P., De~Domenico, M.: Assessing
  the risks of `infodemics' in response to {{COVID}}-19 epidemics. Nature Human
  Behaviour  \textbf{4}(12),  1285--1293 (Dec 2020)

\bibitem{giachanouImpactEmotionalSignals2021}
Giachanou, A., Rosso, P., Crestani, F.: The impact of emotional signals on
  credibility assessment. Journal of the Association for Information Science
  and Technology  \textbf{72}(9),  1117--1132 (2021). \doi{10.1002/asi.24480}

\bibitem{gothelfLeanUXDesigning2016}
Gothelf, J., Seiden, J.: Lean {{UX}}: Designing {{Great Products}} with {{Agile
  Teams}}. {O'Reilly Media, Inc.}, {Sebastopol} (Sep 2016)

\bibitem{heCIDOCommunitybasedOntology2020}
He, Y., Yu, H., Ong, E., Wang, Y., Liu, Y., Huffman, A., Huang, H.h., Beverley,
  J., Hur, J., Yang, X., Chen, L., Omenn, G.S., Athey, B., Smith, B.: {{CIDO}},
  a community-based ontology for coronavirus disease knowledge and data
  integration, sharing, and analysis. Scientific Data  \textbf{7}(1), ~181 (Jun
  2020)

\bibitem{hettrickResearchSoftwareSustainability2016}
Hettrick, S.: Research software sustainability: Report on a {{Knowledge
  Exchange}} workshop. Tech. rep., {The Software Sustainability Institute}
  (2016)

\bibitem{houyUnderstandingUnderstandabilityConceptual2012}
Houy, C., Fettke, P., Loos, P.: Understanding understandability of conceptual
  models\textendash what are we actually talking about? In: International
  {{Conference}} on {{Conceptual Modeling}}. pp. 64--77. {Springer}, {Florence,
  Italy} (2012)

\bibitem{jahanbakhshExploringLightweightInterventions2021}
Jahanbakhsh, F., Zhang, A.X., Berinsky, A.J., Pennycook, G., Rand, D.G.,
  Karger, D.R.: Exploring {{Lightweight Interventions}} at {{Posting Time}} to
  {{Reduce}} the {{Sharing}} of {{Misinformation}} on {{Social Media}}.
  Proceedings of the ACM on Human-Computer Interaction  \textbf{5}(CSCW1),
  18:1--18:42 (Apr 2021)

\bibitem{jiang-etal-2020-hover}
Jiang, Y., Bordia, S., Zhong, Z., Dognin, C., Singh, M., Bansal, M.: {{HoVer}}:
  A dataset for many-hop fact extraction and claim verification. In: Findings
  of the Association for Computational Linguistics: {{EMNLP}} 2020. pp.
  3441--3460. {Association for Computational Linguistics}, {Online} (Nov 2020).
  \doi{10.18653/v1/2020.findings-emnlp.309}

\bibitem{juretaComprehensiveQualityModel2009}
Jureta, I.J., Herssens, C., Faulkner, S.: A comprehensive quality model for
  service-oriented systems. Software Quality Journal  \textbf{17}(1),  65--98
  (2009)

\bibitem{kagolovskyNewApproachConcept2001}
Kagolovsky, Y., M{\"o}hr, J.R.: A new approach to the concept of ``relevance''
  in information retrieval ({{IR}}). In: {{MEDINFO}} 2001. pp. 348--352. {IOS
  Press}, {Amsterdam} (2001)

\bibitem{kakolUnderstandingPredictingWeb2017}
Kakol, M., Nielek, R., Wierzbicki, A.: Understanding and predicting {{Web}}
  content credibility using the {{Content Credibility Corpus}}. Information
  Processing \& Management  \textbf{53}(5),  1043--1061 (Sep 2017)

\bibitem{kangQuantifyingPerceivedPolitical2020}
Kang, H., Yang, J.: Quantifying perceived political bias of newspapers through
  a document classification technique. Journal of Quantitative Linguistics
  \textbf{Ahead-of-print}(Ahead-of-print),  1--24 (2020)

\bibitem{karrayHumancomputerInteractionOverview2017}
Karray, F., Alemzadeh, M., Abou~Saleh, J., Arab, M.N.: Human-computer
  interaction: Overview on state of the art. International journal on smart
  sensing and intelligent systems  \textbf{1}(1),  137--159 (2017)

\bibitem{kautzInvestigatingDesignProcess2011}
Kautz, K.: Investigating the design process: Participatory design in agile
  software development. Information Technology \& People  \textbf{24}(3),
  217--235 (Jan 2011)

\bibitem{kellerPoliticalAstroturfingTwitter2020}
Keller, F.B., Schoch, D., Stier, S., Yang, J.: Political astroturfing on
  {{Twitter}}: How to coordinate a disinformation campaign. Political
  Communication  \textbf{37}(2),  256--280 (2020)

\bibitem{kuusinenAgileUserExperience2012}
Kuusinen, K., Mikkonen, T., Pakarinen, S.: Agile user experience development in
  a large software organization: Good expertise but limited impact. In:
  International {{Conference}} on {{Human}}-{{Centred Software Engineering}}.
  pp. 94--111. {Springer}, {Toulouse} (2012)

\bibitem{labropoulouMakingMetadataFit2020a}
Labropoulou, P., Gkirtzou, K., Gavriilidou, M., Deligiannis, M., Galanis, D.,
  Piperidis, S., Rehm, G., Berger, M., Mapelli, V., Rigault, M., Arranz, V.,
  Choukri, K., Backfried, G., {G{\'o}mez-P{\'e}rez}, J.M., {Garcia-Silva}, A.:
  Making {{Metadata Fit}} for {{Next Generation Language Technology
  Platforms}}: The {{Metadata Schema}} of the {{European Language Grid}}. In:
  Proceedings of the 12th {{Language Resources}} and {{Evaluation Conference}}.
  pp. 3428--3437. {European Language Resources Association}, {Marseille,
  France} (May 2020)

\bibitem{leeAgileIntegratedAnalysis2010}
Lee, G., Xia, W.: Toward agile: An integrated analysis of quantitative and
  qualitative field data on software development agility. MIS Quarterly
  \textbf{34}(1),  87--114 (2010)

\bibitem{maChangingConceptsWorking2014}
Ma, W.J., Husain, M., Bays, P.M.: Changing concepts of working memory. Nature
  neuroscience  \textbf{17}(3),  347--356 (Mar 2014). \doi{10.1038/nn.3655}

\bibitem{mackenzieHumanComputerInteractionEmpirical2012}
MacKenzie, I.S.: Human-{{Computer Interaction}}: An {{Empirical Research
  Perspective}}. {Newnes}, {Waltham} (Dec 2012)

\bibitem{mcgrewCanStudentsEvaluate2018}
McGrew, S., Breakstone, J., Ortega, T., Smith, M., Wineburg, S.: Can students
  evaluate online sources? learning from assessments of civic online reasoning.
  Theory \& Research in Social Education  \textbf{46}(2),  165--193 (2018)

\bibitem{michener2013identifying}
Michener, G., Bersch, K.: Identifying transparency. Information Polity
  \textbf{18}(3),  233--242 (2013)

\bibitem{nielsen1994estimating}
Nielsen, J.: Estimating the number of subjects needed for a thinking aloud
  test. International journal of human-computer studies  \textbf{41}(3),
  385--397 (1994)

\bibitem{ozencHowSupportDesigners2010}
Ozenc, F.K., Kim, M., Zimmerman, J., Oney, S., Myers, B.: How to support
  designers in getting hold of the immaterial material of software. In:
  Proceedings of the {{SIGCHI Conference}} on {{Human Factors}} in {{Computing
  Systems}}. pp. 2513--2522. {ACM}, {Atlanta} (2010)

\bibitem{pankovskaSuspiciousSentenceDetection2022}
Pankovska, E., Schulz, K., Rehm, G.: Suspicious {{Sentence Detection}} and
  {{Claim Verification}} in the {{COVID-19 Domain}}. In: Proceedings of the
  Workshop {{Reducing Online Misinformation}} through {{Credible Information
  Retrieval}} ({{ROMCIR}} 2022). {CEUR-WS}, {Stavanger} (2022)

\bibitem{pasiDecisionMakingMultiple2020}
Pasi, G., De~Grandis, M., Viviani, M.: Decision making over multiple criteria
  to assess news credibility in microblogging sites. In: 2020 {{IEEE
  International Conference}} on {{Fuzzy Systems}} ({{FUZZ}}-{{IEEE}}).
  pp.~1--8. {IEEE}, {Glasgow} (2020)

\bibitem{patwaFightingInfodemicCOVID192021}
Patwa, P., Sharma, S., Pykl, S., Guptha, V., Kumari, G., Akhtar, M.S., Ekbal,
  A., Das, A., Chakraborty, T.: Fighting an {{Infodemic}}: {{COVID}}-19 {{Fake
  News Dataset}}. arXiv:2011.03327 [cs]  \textbf{1402},  21--29 (2021)

\bibitem{przybylaWhenClassificationAccuracy2021}
Przyby{\l}a, P., Soto, A.J.: When classification accuracy is not enough:
  Explaining news credibility assessment. Information Processing \& Management
  \textbf{58}(5),  102653 (Sep 2021)

\bibitem{raisonKeepingUserCentred2013}
Raison, C., Schmidt, S.: Keeping user centred design ({{UCD}}) alive and well
  in your organisation: Taking an agile approach. In: International
  {{Conference}} of {{Design}}, {{User Experience}}, and {{Usability}}. pp.
  573--582. {Springer}, {Las Vegas} (2013)

\bibitem{rehmInfrastructureEmpoweringInternet2018}
Rehm, G.: An {{Infrastructure}} for {{Empowering Internet Users}} to {{Handle
  Fake News}} and {{Other Online Media Phenomena}}. In: Rehm, G., Declerck, T.
  (eds.) Language {{Technologies}} for the {{Challenges}} of the {{Digital
  Age}}. pp. 216--231. Lecture {{Notes}} in {{Computer Science}}, {Springer
  International Publishing}, {Cham} (2018)

\bibitem{rehmEuropeanLanguageGrid2020}
Rehm, G., Berger, M., Elsholz, E., Hegele, S., Kintzel, F., Marheinecke, K.,
  Piperidis, S., Deligiannis, M., Galanis, D., Gkirtzou, K., Labropoulou, P.,
  Bontcheva, K., Jones, D., Roberts, I., Haji{\v c}, J., Hamrlov{\'a}, J.,
  Ka{\v c}ena, L., Choukri, K., Arranz, V., Vasi{\c l}jevs, A., Anvari, O.,
  Lagzdi{\c n}{\v s}, A., Me{\c l}{\c n}ika, J., Backfried, G., Dikici, E.,
  Janosik, M., Prinz, K., Prinz, C., Stampler, S., {Thomas-Aniola}, D.,
  {G{\'o}mez-P{\'e}rez}, J.M., Garcia~Silva, A., Berr{\'i}o, C., Germann, U.,
  Renals, S., Klejch, O.: European {{Language Grid}}: {{An Overview}}. In:
  Proceedings of the 12th {{Language Resources}} and {{Evaluation Conference}}.
  pp. 3366--3380. {European Language Resources Association}, {Marseille,
  France} (May 2020)

\bibitem{rehmEuropeanLanguageGrid2021}
Rehm, G., Piperidis, S., Bontcheva, K., Hajic, J., Arranz, V., Vasi{\c l}jevs,
  A., Backfried, G., {G{\'o}mez-P{\'e}rez}, J.M., Germann, U., Calizzano, R.:
  European language grid: {{A}} joint platform for the european language
  technology community. In: Proceedings of the 16th {{Conference}} of the
  {{European Chapter}} of the {{Association}} for {{Computational
  Linguistics}}: {{System Demonstrations}}. pp. 221--230 (2021)

\bibitem{rehm2018c}
Rehm, G., Schneider, J.M., Bourgonje, P.: Automatic and manual web annotations
  in an infrastructure to handle fake news and other online media phenomena.
  In: Calzolari, N., Choukri, K., Cieri, C., Declerck, T., Goggi, S., Hasida,
  K., Isahara, H., Maegaard, B., Mariani, J., Mazo, H., Moreno, A., Odijk, J.,
  Piperidis, S., Tokunaga, T. (eds.) Proceedings of the 11th Language Resources
  and Evaluation Conference ({{LREC}} 2018). pp. 2416--2422. {European Language
  Resources Association (ELRA)}, {Miyazaki, Japan} (May 2018)

\bibitem{riegerCorona100dGermanlanguageTwitter2021}
Rieger, J., {von Nordheim}, G.: Corona100d: German-language {{Twitter}} dataset
  of the first 100 days after {{Chancellor Merkel}} addressed the coronavirus
  outbreak on {{TV}}. Tech. rep., {DoCMA Working Paper} (2021)

\bibitem{riegerGermanChineseDataset2020}
Rieger, M.O., {He-Ulbricht}, Y.: German and {{Chinese}} dataset on attitudes
  regarding {{COVID}}-19 policies, perception of the crisis, and belief in
  conspiracy theories. Data in Brief  \textbf{33},  106384 (Dec 2020)

\bibitem{riehCredibilityAssessmentOnline2014}
Rieh, S.Y.: Credibility assessment of online information in context. Journal of
  Information Science Theory and Practice  \textbf{2}(3),  6--17 (2014)

\bibitem{rogersQADatasetExplosion2021}
Rogers, A., Gardner, M., Augenstein, I.: {{QA Dataset Explosion}}: A
  {{Taxonomy}} of {{NLP Resources}} for {{Question Answering}} and {{Reading
  Comprehension}}. arXiv:2107.12708 [cs]  \textbf{2107}(12708),  1--38 (Jul
  2021)

\bibitem{saltzMisinformationInterventionsAre2021}
Saltz, E., Barari, S., Leibowicz, C., Wardle, C.: Misinformation interventions
  are common, divisive, and poorly understood. Harvard Kennedy School
  Misinformation Review  \textbf{2}(5),  1--25 (Oct 2021).
  \doi{10.37016/mr-2020-81}

\bibitem{samimiDeclarativeMocking2013}
Samimi, H., Hicks, R., Fogel, A., Millstein, T.: Declarative mocking. In:
  Proceedings of the 2013 {{International Symposium}} on {{Software Testing}}
  and {{Analysis}}. pp. 246--256. {ACM}, {New York, NY} (2013)

\bibitem{sassGermanCoronaConsensus2020}
Sass, J., Bartschke, A., Lehne, M., Essenwanger, A., Rinaldi, E., Rudolph, S.,
  Heitmann, K.U., Vehreschild, J.J., {von Kalle}, C., Thun, S.: The {{German
  Corona Consensus Dataset}} ({{GECCO}}): A standardized dataset for
  {{COVID}}-19 research in university medicine and beyond. BMC Medical
  Informatics and Decision Making  \textbf{20}(1), ~341 (Dec 2020)

\bibitem{sauroQuantifyingUserExperience2016}
Sauro, J., Lewis, J.R.: Quantifying the User Experience: Practical Statistics
  for User Research. {Morgan Kaufmann}, {Cambridge, MA} (2016)

\bibitem{solisStudyCharacteristicsBehaviour2011}
Solis, C., Wang, X.: A study of the characteristics of behaviour driven
  development. In: Proceedings of the 37th {{EUROMICRO Conference}} on
  {{Software Engineering}} and {{Advanced Application}}. pp. 383--387. {IEEE},
  {Los Alamitos} (2011)

\bibitem{suMotivationsMethodsMetrics2020}
Su, Q., Wan, M., Liu, X., Huang, C.R.: Motivations, {{Methods}} and {{Metrics}}
  of {{Misinformation Detection}}: An {{NLP Perspective}}. Natural Language
  Processing Research  \textbf{1}(1-2),  1--13 (Jun 2020)

\bibitem{teyssouInVIDPluginWeb2017}
Teyssou, D., Leung, J.M., Apostolidis, E., Apostolidis, K., Papadopoulos, S.,
  Zampoglou, M., Papadopoulou, O., Mezaris, V.: The {{InVID Plug}}-in: Web
  {{Video Verification}} on the {{Browser}}. In: Proceedings of the {{First
  International Workshop}} on {{Multimedia Verification}}. pp. 23--30.
  {{MuVer}} '17, {Association for Computing Machinery}, {New York, NY, USA}
  (Oct 2017)

\bibitem{thakurBEIRHeterogeneousBenchmark2021}
Thakur, N., Reimers, N., R{\"u}ckl{\'e}, A., Srivastava, A., Gurevych, I.:
  {{BEIR}}: A {{Heterogeneous Benchmark}} for {{Zero}}-shot {{Evaluation}} of
  {{Information Retrieval Models}}. In: Thirty-Fifth {{Conference}} on {{Neural
  Information Processing Systems Datasets}} and {{Benchmarks Track}} ({{Round}}
  2). pp. 1--16. {NeurIPS}, {Virtual} (Aug 2021)

\bibitem{tuTransparencySoftwareEngineering2014}
Tu, Y.C.: Transparency in Software Engineering. Ph.D. thesis, The University of
  Auckland, {Auckland} (2014)

\bibitem{tu2016experiment}
Tu, Y.C., Tempero, E., Thomborson, C.: An experiment on the impact of
  transparency on the effectiveness of requirements documents. Empirical
  Software Engineering  \textbf{21}(3),  1035--1066 (2016)

\bibitem{vargasDetectionDisinformationCampaign2020}
Vargas, L., Emami, P., Traynor, P.: On the detection of disinformation campaign
  activity with network analysis. In: Proceedings of the 2020 {{ACM SIGSAC
  Conference}} on {{Cloud Computing Security Workshop}}. pp. 133--146. {ACM},
  {Virtual} (2020)

\bibitem{vivianiCredibilitySocialMedia2017}
Viviani, M., Pasi, G.: Credibility in social media: Opinions, news, and health
  information\textemdash a survey. Wiley interdisciplinary reviews: Data mining
  and knowledge discovery  \textbf{7}(5),  e1209 (2017)

\bibitem{wauteletUnifyingExtendingUser2014}
Wautelet, Y., Heng, S., Kolp, M., Mirbel, I.: Unifying and extending user story
  models. In: International {{Conference}} on {{Advanced Information Systems
  Engineering}}. pp. 211--225. {Springer}, {Thessaloniki} (2014)

\bibitem{williams1977experimental}
Williams, E.: Experimental comparisons of face-to-face and mediated
  communication: A review. Psychological Bulletin  \textbf{84}(5), ~963 (1977)

\bibitem{wobbrockGoldilocksZoneYoung2021}
Wobbrock, J.O., Hattatoglu, L., Hsu, A.K., Burger, M.A., Magee, M.J.: The
  {{Goldilocks}} zone: Young adults' credibility perceptions of online news
  articles based on visual appearance. New Review of Hypermedia and Multimedia
  \textbf{0}(0),  1--46 (Feb 2021)

\bibitem{zhouReCOVeryMultimodalRepository2020}
Zhou, X., Mulay, A., Ferrara, E., Zafarani, R.: {{ReCOVery}}: A {{Multimodal
  Repository}} for {{COVID}}-19 {{News Credibility Research}}. In: Proceedings
  of the 29th {{ACM International Conference}} on {{Information}} \&
  {{Knowledge Management}}. pp. 3205--3212. {ACM}, {Virtual Event Ireland} (Oct
  2020)

\end{thebibliography}
\appendix
\section{Newspaper Article (Translation)}
\label{appendix:newspaperArticle}
Spanish youths receive cultural vouchers

The culture industry around the world suffered from the corona pandemic. In Spain, young people now receive EUR 400 vouchers to take advantage of cultural offers - but one type of event is excluded. 

To cushion the hardships of the corona pandemic, young people in Spain receive a cultural voucher . Everyone who will turn 18 next year should receive a voucher worth 400 euros from the government. But it cannot be used indefinitely: the recipients cannot buy tickets for bullfights with it.

The decision was one of the politically controversial measures the government included in the 2022 draft state budget. The vouchers are intended to help the country's culture and events industry recover from the loss of income during the corona lockdowns. According to the government, eligible teenagers can spend their 400 euros on cinema and theater tickets, books and concerts, for example.

The Ministry of Culture announced in a written communication to the state news agency Efe that ``not everything that our legislation regards as culture will fall under this cultural support.'' Bullfighting is now rejected by a large part of Spanish society, especially young city dwellers.

\end{document}